# A Model for Combination of External and Internal Stimuli in the Action Selection of an Autonomous Agent

Pedro Pablo González Pérez[1]
Instituto de Investigaciones Biomédicas/UNAM
Instituto de Química/UNAM
ppgp@servidor.unam.mx

José Negrete Martínez
Instituto de Investigaciones Biomédicas/UNAM
Maestría en Inteligencia Artificial/UV
jnegrete@mia.uv.mx

Ariel Barreiro García
Centro de Cibernética Aplicada a la Medicina/ISCM-H
ariel@cecam.cu

Carlos Gershenson García
Instituto de Química/UNAM
Fundación Arturo Rosenblueth
Facultad de Filosofía y Letras/UNAM
carlos@jlagunez.iquimica.unam.mx

*Abstract*

*This paper proposes a model for combination of external and internal stimuli for the action selection in an autonomous agent, based in an action selection mechanism previously proposed by the authors. This combination model includes additive and multiplicative elements, which allows to incorporate new properties, which enhance the action selection. A given parameter $\alpha$, which is part of the proposed model, allows to regulate the degree of dependence of the observed external behaviour from the internal states of the entity.*

## 1. Introduction

The theory of behaviour-based systems (Brooks, 1986; Brooks, 1989) provides a new philosophy for the construction of autonomous agents, inspired in the ethology field. Unlike the knowledge-based systems, behaviour-based systems interact directly with their problem domain. An autonomous agent perceives its problem domain through its sensors and acts on this through its actuators. The problem domain of an autonomous agent is commonly a dynamic, complex and unpredictable environment, in which it deals to satisfy a set of goals or motivations, which could vary in time. An autonomous agent decides itself how to relate its external and internal stimuli to its motor actions in such a way that its goals can be satisfied (Maes, 1994).

---

[1] To whom all correspondence should be sent: Laboratorio de Qu mica Te rica. Instituto de Qu mica. Circuito Universitario, Coyoac n, M xico C.P. 04510 tel. (52) 5 622 4424

The term "agent" has been widely used in a great variety of research domains, being the most common those domains belonging to artificial intelligence and computer science areas. Several have been the definitions and interpretations contributed for the term "agent", depending these on the domain in which the agent has been used, as well as of the particular use that it has been given.

According to Maes (Maes, 1994) " an agent is a system that tries to satisfy a set of goals in a dynamic and complex environment. An agent is situated in the environment: it can sense the environment through its sensors and act upon the environment using its actuators".

An agent has goals, and in dependency of the domain in which the agent has been defined, these can take different forms. In order to satisfy its goals, an autonomous agent must select, at every moment in time, the most appropriate action among all the possible actions that it could execute. An action selection mechanism (ASM) is a computational mechanism, which must produce a selected action as output when different external or internal stimuli have been provided as inputs. That is, an ASM decides how to combine external and internal stimuli to select which action must be executed by the agent.

In this paper we propose a model for combination of external and internal stimuli, from an ASM developed by the authors (González, 1999), for the action selection in an autonomous agent, which exhibits some of the properties not satisfied or not explained by the ASMs that will be reviewed here. The proposed model incorporates new properties derived from the principles that characterize the animal behaviour, which must enrich the action selection.

The ASM has been structured in a distributed blackboard architecture, which, given its great capacity for the coordination and integration of several tasks in real time and its extreme flexibility for the incorporation of new functionality, facilitates the model implementation. As well as, the incremental incorporation of new properties and processes of learning directed to enrich the selection of actions, making it more adaptive.

The paper is structured as follows. The next section briefly describes some of main ASMs proposed in the literature and a comparison between these ASMs is done. This comparison is focussed to the combination scheme utilized. Section 3 presents the ASM developed by the authors, the Internal Behaviour Network (González, 1999). Here are discussed the essential characteristics of the proposed ASM that allows to understand the combination scheme that will be presented in section 4. Section 5 presents the simulation and experiments developed to verify the properties of the combination scheme.

## 2. Action Selection Mechanisms

The following are examples of ASMs and related works to action selection:

- Tinbergen's mechanism (Tinbergen, 1950; Tinbergen, 1951), a hierarchic network of nodes or centres, which approaches the complete action selection problem with a noticeable emphasis in the reproductive stage.
- Lorenz's psycho-hydraulic mechanism (Lorenz, 1950; Lorenz, 1981), a model that tries to explain some ethology phenomena, without approaching the action selection problem complete.
- Baerends' model (Baerends, 1976), a hierarchic network of nodes, a model

- inspired by ethologist studies made in particular classes of insects and birds.
- Brooks' subsumption architecture (Brooks, 1986; Brooks, 1989), which can be used as a mean to implement robot control systems, which include tasks of perception and action, in addition to the emergency of behaviours
- Rosenblatt and Payton's hierarchical network (Rosenblatt and Payton, 1989), a mechanism very similar in many aspects to the hierarchical models proposed by Tinbergen and Baerends, but with nodes very similar to formal neurons.
- Maes' bottom-up mechanism (Maes, 1990; Maes, 1991), a distributed non-hierarchical network of nodes, where each node represents an appetitive or consummatory alternative that the entity can execute.
- Beer's neural model (Beer, 1990; Beer, Chiel and Sterling, 1990), a semi-hierarchical network of nodes, where each node represents a neuronal circuit.
- Halperin's neuroconnector network (Hallam, Halperin and Hallam, 1994), a non-supervised neural network organized in layers.
- Negrete's neuro-humoral model (Negrete and Martinez, 1996), a non-hierarchical distributed network of nodes, where each node is a neuro-humoral neuron.
- Goetz's recurrent behaviour network (Goetz and Walters, 1997), a network of nodes, where a node can represent a behaviour, a sensor or a goal; the network converges to a particular behaviour (attractor), in a similar way that a Hopfield's network (Hopfield, 1982) converges to a certain pattern.

Table 1 shows a comparison between different ASMs before mentioned, taking into account the most relevant aspects from these. In particular, our interest is focussed to the form in which these mechanisms combine the external and internal stimuli to select an external action.

| ASM | Disciplines | Architecture | Combination of stimuli | Learning schemes |
|---|---|---|---|---|
| Tinbergen | ethology | hierarchical network of nodes, where each node represents a kind of behaviour | summed | none |
| Lorenz | ethology, psychology and hydraulic engineering | psycho-hydraulic model | summed | none |
| Baerends | ethology | hierarchical network of nodes, where each node represents a kind of behaviour | unstated | none |
| Brooks | robotic | distributed network of finite state machines | unstated | |
| Rosenblatt and Payton | robotic and artificial neural networks | connectionist, feed-forward network, behaviours are defined by connections between processing elements | can be any function of weighted inputs | none |

| Maes | ethology and behaviour-based systems | non-hierarchical, distributed network, where each node represents a kind of behaviour | summed | none |
|---|---|---|---|---|
| Beer | ethology, neuroethology and artificial neural network | semi-hierarchical network, where each node is a neural network implementing a particular kind of behaviour | summed | none |
| Halperin | ethology and artificial neural network | non-supervised, hierarchical, feed-forward network | summed | classical, secondary, and postponed conditioning |
| Negrete | neurophysiology ethology | non-hierarchical, distributed network of neuro-humoral neurons | summed | none |
| Goetz | artificial neural network and attractors theory | recurrent distributed network | summed | none |

Table 1.

The great majority of these action selection mechanisms propose an additive model for stimuli combination, as showed in Table 1. The stimuli addition could be interpreted in many ways, but the essence is that this type of stimuli combination models a behaviour with a more reactive than a motivated tendency. This can be explained in the following terms: for high values of the external inputs, the additive combination could surpass some value threshold previously established and, therefore, shoot a certain external action, still for non significant values of the internal inputs.

On the other hand, in a pure model of multiplicative combination both types of inputs are necessary so that an external consummatory action of the entity can be executed. This is, if we considered that $\Sigma_j S_j$ is the sum of all the associated external inputs to the internal state $E_i$, then $\Sigma_j S_j * E_i \neq 0$ only if $\Sigma_j S_j \neq 0$ and $E_i \neq 0$. Pure multiplicative models have a limit ratio to very small values of the external inputs and/or of the internal inputs, since it does not matter how great is the external input if the internal input is near zero and vice versa. A more detailed discussion of the advantages and disadvantages of the ASMs here mentioned can be found in (González, 1999).

An alternative to eliminate the undesired effects of the pure additives and multiplicative combinations is to consider a model that incorporates the most valuable elements of both types of combination. In this paper, we propose a model of external and internal inputs combination for the action selection, where both types of combinations take place. This is, in the proposed model the external and internal inputs interact multiplicatively, which can be interpreted considering that both factors are necessary to activate an external consummatory action. Nevertheless, the model also contains additive elements, which allow to give a greater importance or weight to the internal inputs, and are finally those that determine that interesting properties as the action selection directed towards internal medium of the entity, and the execution of an external behaviour oriented to the search of a specific signal can emerge product from this type of combination.

## 3. The Internal Behaviour Network

The ASM proposed by us has been structured from a network of blackboard nodes developed by the authors (González and Negrete, 1997; Negrete and González, 1998). A blackboard node is integrated by the following components: a set of independent modules called knowledge sources, which have specific knowledge about the problem domain; the blackboard, a shared data structure through which the knowledge sources communicate to each other by means of the creation of solution elements on the blackboard; the communication mechanisms, which establish the interface between the nodes and a node and the external and internal mediums; the activation state registers of the knowledge sources (REAC); and a control mechanism, which determines the order in that the knowledge sources will operate on the blackboard.

As it can be appreciated in Figure 1, the actual architecture of the ASM exhibits two blackboard nodes: the cognitive node and the motivational node. This architecture has been called Internal Behaviour Network (IBeNet) (González, 1999). The term " internal behaviour " has been used to describe the processes of creation and modification of solution elements that occur on the blackboard of the node. Structurally, an internal behaviour is a package of production rules (if <condition> then <action>). A production rule is also known as an elementary behaviour. The condition of an elementary behaviour describes the configuration of solution elements on the blackboard that is necessary, so that the elementary behaviour contributes to the solution processes of the problem. The way in which an elementary behaviour contributes to the solution of the problem is specified in its action, which can consist of the creation or modification of solution elements in certain blackboard levels.

The cognitive node blackboard is structured in six levels of abstraction: external perceptions, perceptual persistents, consummatory preferents, drive/perception congruents, potential actions, and actions. For the cognitive node, the following internal behaviours have been defined: perceptual persistence, attention to preferences, reflex response inhibition, and external behaviours selector. The defined mechanisms of communication for this node are the exteroceptors and actuators, which establish the interface between the cognitive node and external medium; and the receptor and transmitter mechanisms, which establish communication with the motivational node.

The role of the cognitive node includes the processes of representation of the perceptual signals, integration of internal and external signals, inhibition of reflex responses, and the selection of the external behaviour that adjusts better to the actual external conditions and internal needs.

The blackboard of the motivational node is structured in four abstraction levels: internal perceptions, external perceptions, propio/extero/drive congruents, and drive. The internal behaviours that operate in the motivational node are the following: propio/extero/drive congruence and consummatory preferences selector. The communication of this node is established from the propioceptors, which define the interface between the node and the internal medium; and the receptor and transmitter mechanisms, which establish the communication with the cognitive node.

The role of the motivational node includes the processes of representation of internal signals, combination of internal and external signals and the selection of the more appropriate consummatory preference.

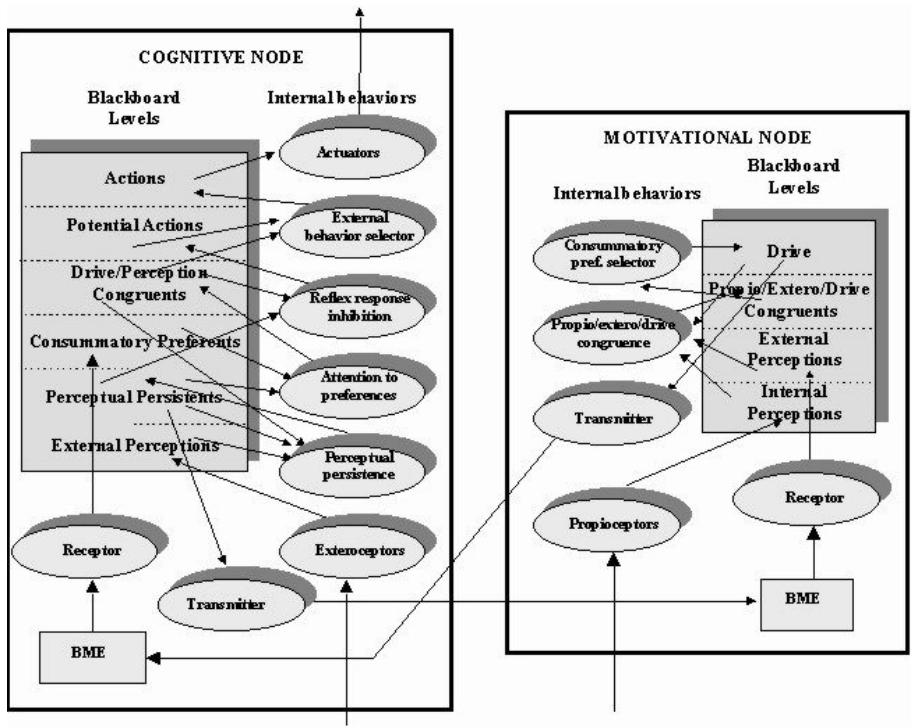

Figure 1. Structure of the blackboard node network. Here have been omitted the activity state registers.

A full presentation of the structure and properties of the IBeNet can be found in (González, 1999).

## 4. The Model fo External and Internal Stimuli Combination

At the level of the propio/extero/drive congruence internal behaviour of the motivational node is carried out the external and internal signals combination to determine the action that must be executed by the entity. These signals are registered in the external perceptions and internal perceptions levels, respectively, of the motivational node. The solution elements registered in the external perceptions level are originated by unconditional stimuli (US), conditional stimuli (CS), or both; all of them initially projected in the external perceptions level and later recovered from the perceptual persistents level, of the cognitive node. On the other hand, the solution elements registered in the internal perceptions level of the motivational node represent the values of the internal states that are able to satisfy part of the conditions of certain propio/extero/drive congruence elementary behaviours that operate on this level. The scheme of signals combination used by this internal behaviour is the one given by the expression (1).

$$A_i^C = O_i^E * (\alpha + \Sigma_j Fa_{ij}^S * O_j^S) + O_i^D \tag{1}$$

where: $A_i^C$ is the certainty value with which will be created the element solution $C_i$ in the propio/extero/drive congruents level of the blackboard, $\alpha$ is the weight or importance attributable to the internal state ($0 \pounds \alpha \pounds 1$), $O_i^E$ is the internal signal, $O_i^D$ is the signal created in the Drive level ($O_i^D = 0$, for all i at the initial moment $t = t_0$), $O_j^S$ are the external signals associated to the internal state $O_i^E$, and $Fa_{ij}^S$ are the coupling strengths of the elemental behaviours propio/extero/drive congruence (coupling strengths are to elemental behaviours what synaptic weights are to artificial neural networks).

The propio/extero/drive congruence internal behaviour does not require the creation of REACs. Whenever the condition of an elementary behaviour is satisfied, this executes its final action, creating the solution element $C_i$ with certainty $A_i^C$ in the propio/extero/drive congruents level of the blackboard of the motivational node. The level of certainty $A_i^C$ represents the activity of the elementary behaviour i. Figure 2 shows the structure of an elementary behaviour of this type.

**Parameters:**
    condition solution elements: $E_i$, $S_j$, $D_i$ (j = 1, 2, ...)
    action solution elements: $C_i$
    coupling strengths: $Fa_{ij}^S$ (j = 1, 2, ...)
    importance attributable to the internal state: $\alpha$

**Condition:**
    ($\alpha = 0$ AND
    blackboard-level = Internal Perceptions AND
    cert($E_i$, $O_i^E$) AND
    $O_i^E \neq 0$ AND
    blackboard-level = External Perceptions AND
    cert($S_i$, $O_i^S$) AND
    $O_i^S \neq 0$) OR
    ($\alpha \neq 0$ AND
    blackboard-level = Internal Perceptions AND
    cert($E_i$, $O_i^E$) AND
    $O_i^E \neq 0$ AND
    blackboard-level = External Perceptions AND
    cert($S_i$, $O_i^S$))

**Action:**
    calculate $A_i^C = O_i^E * (\alpha + \Sigma_j Fa_{ij}^S * O_j^S) + O_i^D$ AND
    blackboard-level = Propio/Extero/Drive Congruents AND
    create-solution-element ($C_i$, $A_i^C$)

Figure 2. Structure of a propio/extero/drive congruence elementary behaviour.

The role of the propio/extero/drive congruence elementary behaviour are next described for the cases when $\alpha = 1$ and $\alpha \neq 0$.

**Case 1: $\alpha=0$**

When $\alpha$ is equal to zero, the propio/extero/drive congruence elementary behaviour will be able to activate only when the internal and external signals ($O_i^E \neq 0$ and $\Sigma_j Fa_{ij}^S * O_j^S \neq 0$) associated to this coincide. If the necessary internal state is not present, then the external signal is not sufficient by itself to evoke the motor action, since the attention to preferences internal behaviour is activated only when their external signal (solution element in the perceptual persistent level of the cognitive node) and the drive signal (solution element in the consummatory preferents of the cognitive node) coincide. On the other hand, although a very strong internal signal exists, corresponding to an urgent necessity of the entity, if the usable external signal is not present, the propio/extero/drive congruence elementary behaviour will not activate. Then, when $\alpha = 0$, both factors (external signal and internal signal) are necessary so that the propio/extero/drive congruence internal behaviour activates. Physiologically this can be interpreted, when thinking that the inputs that come from the internal states sensitize this internal behaviour with the signals that come from the external medium.

**Case 2: $\alpha \neq 0$**

When $\alpha$ is different of zero, then there is being granted a little more weight (or importance) to the internal state than to the external inputs. So that, still in total absence of external inputs, the propio/extero/drive congruence elementary behaviour could be activated for a very strong value of the internal signals (when the value of $\alpha$ is near one). Thus, when the internal necessity is extremely high and external signals do not exist, the propio/extero/drive congruence elementary behaviour can be activated, and therefore to cause a type of "preactivation" of the conditions of the attention to preferences elementary behaviours (in particular, only one of these elementary behaviours will be the one that will receive greater preactivation). This it is a typical example of an internal behaviour whose actions are directed to the internal medium. This mechanism constitutes the underlying base of the exploratory behaviour oriented to a specific objective, which is another of the distinguishing properties of the IBeNet. When there is a total satiety of the internal necessity that caused the activation of the propio/extero/drive congruence elementary behaviour ($O_i^E = 0$), then the level of activation of this elementary behaviour will decrease to zero ($A_i^C = 0$), independently of the existence of usable external signals. For activation levels of zero or very near to zero the elementary behaviour will not be activated. Of this way, the observed external behaviour always will be adaptive.

## 5. Simulation and Experiments

### 5.1. The simulation

In order to verify the role of the parameter $\alpha$ in the combination of external and internal inputs, a program was written implementing the IbeNet, and used by an autonomous mobile robot (animat) simulated in virtual reality. The simulated robot was inspired by an original idea developed by Negrete (Negrete and Martinez, 1996). The graphical representation of the animat is a cone with a semicircumscript sphere. This has primitive motor and perceptual systems developed specially for this simulation. The animat has internal states to satisfy, such as hunger, thirst and fatigue, and the qualities of strength and lucidity; which are affected by the diverse behaviours that this can execute. The implemented behaviours were: wander, explore, approach a specific signal, avoid obstacles, rest, eat, drink, and runaway. An environment was also simulated, in which the animat develops. In this environment, obstacles, sources of food and water, grass, blobs (aversive stimuli), and spots of different magnitudes can be created, which animat can use to satisfy its internal necessities. This simulation can be accessed via Internet in the address http://132.248.11.4/~carlos/asia/animat/animat.html .

**5.2. The experiments**

The developed experiments using this simulation were directed to verify: (1) the influence of the internal states in the observed external behaviour of the animat, (2) the role of the competition at a motivational level in the selection of the external behaviour to execute, (3) the exploratory behaviour oriented to the search of a specific signal and the reduction of the response times of the animat, (4) the stability in the selection and persistence in the execution of the external behaviours, (5) the discrimination between different stimuli taking in count the quality of them, (6) avoid aversive stimuli, (7) the non-persistence in the execution of a consummatory action when an aversive stimulus is perceived, and (8) the role of the learning processes in the action selection. Here, experiment (3) is illustrated, which was oriented to the verification of the following hypothesis:

"The value of $\alpha$ in the expression (1) is closely related with the time of response of the animat between the presentation of an external stimulus, for which a very high internal necessity exists, and the consummation of the action associated to the satisfaction of this necessity, being this an inversely proportional relation".

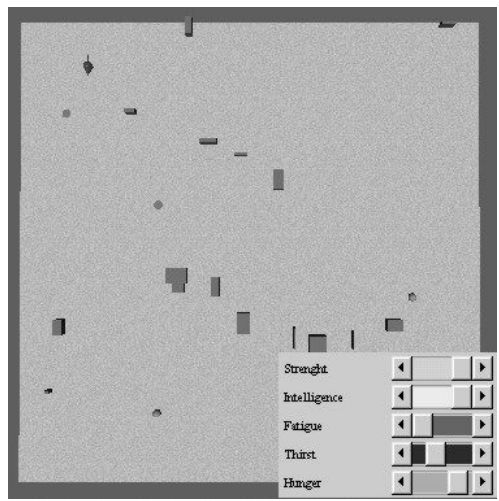

Figure 3. Initial state: water perceived, food not perceived, little thirst and much hunger.

That is, if the value of α increases, then the time of reaction decreases; and if the value of α diminishes, then the time of reaction increases.

In order to demonstrate this hypothesis, we have created a variable that quantifies the time of reaction of the animat, from the moment at which the stimulus in the external medium appeared until the moment at which the corresponding action is executed. The value of this variable, which we will denote like RTIME, is calculated in the following way: given an external stimulus $O_i^+$, for which a very high internal necessity exists, the value of RTIME is equal to the amount of passed cycles of execution from the moment at which the $O_i^+$ stimulus is presented, until the moment at which the corresponding action is executed.

Now let us consider a simulated environment as the one shown in Figure 3, in which the stimulus "food source" is not perceived; whereas several stimuli "water source" can be perceived. Let us consider in addition that exists an urgent "eat" need, whereas the "drink" need is not relevant, as it is reflected in the values of the internal states in this same figure. Having in count these considerations, we analyse now which will be the action executed by animat in each one of the cases α = 0,0 and α ≈ 1.0.

**Case 1: α = 0**

Let us remember that when α is equal to zero, a propio/extero/drive congruence elementary behaviour will be able to be activated only when the external and internal signals associated to it agree. Then, in this case propio/extero/drive congruence elementary behaviour associated to thirst internal state will be activated, since for this there is an usable external input and an internal state, that although is not high, it is different from zero; whereas the propio/extero/drive congruence elementary behaviour associated to the hunger internal state cannot be activated, since although the value of this internal state very is elevated, there is no external signal able to satisfy this (we remember that the food has not been perceived).

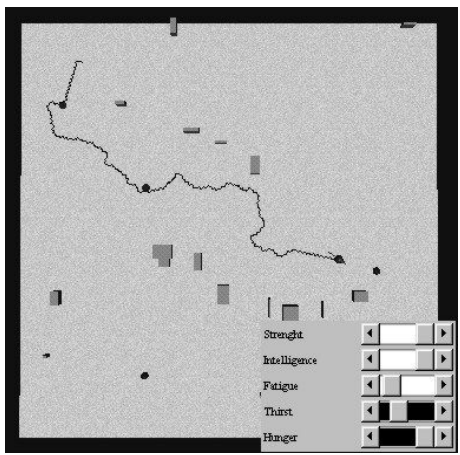

Figure 4. Trajectory and actions executed by the animat, considering the shown initial state in 3, and α = 0.

By virtue of the previous, and as shown in Figure 4, what has happened is that the propio/extero/drive congruence elementary behaviour associated to the thirst internal input prevailed at a motivational level, reason why the drink external behaviour was executed, and the animat has lost time in finding the food, whereas the hunger internal state continues to increase.

**Case 2:** $\alpha \approx 1$

Here we must consider that when $\alpha \approx 1$, to the internal state associated to a propio/extero/drive congruence elementary behaviour is being granted with a little more weight than to the external inputs associated to this, so that, still in total absence of external inputs, this elementary behaviour could be activated for a very strong value of the internal state. Therefore, in this case, both the propio/extero/drive congruence elementary behaviour associated to the thirst internal state, and the one associated to the hunger internal state will be activated, reason why in the competitive process that is carried out to motivational level both elementary behaviours will participate.

By virtue of the previous, and as it can be seen in Figure 5, for a very strong value of $\alpha$ (near 1), the propio/extero/drive congruence elementary behaviour associated to the hunger internal input was winner in the competition, given the low value of the thirst internal input. This fact caused the "preactivation" of the conditions of the attention to preferences elementary behaviours that operate on the blackboard of the cognitive node, with none of these being activated. Because of the previous, the external behaviour exploration oriented to the search of a necessary external signal was activated, being in this case "food source" such signal.

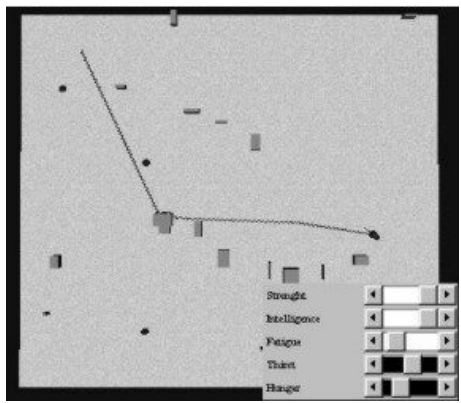

Figure 5. Trajectory and actions executed by animat, considering the initial state shown in Figure 3, and $\alpha \approx 1$.

Furthermore, when the exploratory behaviour oriented to the search of a specific signal is being generated, the animat has a higher probability of finding the source of food faster that in the previous case, when $\alpha$ was equal to 0; which is more adaptive, since the animat's urgent necessity is to eat. To this, it is necessary to add that the response time between the moment at which the food source is perceived directly and the moment in which the eat behaviour is executed will be much smaller, although there are other internal inputs present.

## 6. Conclusions

One of the main results derived from the proposed combination scheme of external and internal inputs, has been to understand that the essential difference between the diverse ASMs reviewed goes beyond the type of architecture in which these have been structured, or of the model by which these have been inspired. The essential difference is in the way in which these ASMs combine the external and internal stimuli to select a determined action, and therefore, in the diverse repertoire of properties that characterizes this combination.

In this way, some of the properties observed in the developed actions selection mechanism are derived directly from the value of $\alpha$ in the combination scheme. Examples of these are: the strong dependency of the observed external behaviour of the internal states of the entity, and the actions selection towards external medium and towards the internal medium of the entity. Among other properties of interest exhibited by the mechanism, there are: the stability in the selection of actions, the persistence in the execution of an action, the existence of an external behaviour oriented to the search of a specific signal, and the explicit relation of the selection of actions with the learning processes.

Based on the characteristics already exposed, the degree of dependency of the observed external behaviour of the internal states of the entity can be regulated with the parameter $\alpha$. Of this way, when $\alpha$ is near zero, the observed behaviour will have a tendency to the reactivity; whereas when $\alpha$ is near one, the observed behaviour will be motivated. Depending on the problem domain in which the agent is implementing the IBeNet, the parameter $\alpha$ grants flexibility to determine the type of relation to establish between the internal and external inputs, giving a greater or smaller importance to these.

Following this idea, considering a dynamic value of $\alpha$, which is learned by the agent, and establishing different values of $\alpha$ for each behaviour; would allow the agent to adapt in order to decide in what situations it must behave in a more motivated or more reactive way. The development of this idea is part of the future work of the authors.